\documentclass[10pt,twocolumn,letterpaper]{article}

\usepackage[pagebackref=true,breaklinks=true,colorlinks,bookmarks=false]{hyperref}
\usepackage{iccv}
\usepackage{times}
\usepackage{epsfig}
\usepackage{graphicx}
\usepackage{amsmath}
\usepackage{amssymb}
\usepackage{tabularx}

\usepackage{eso-pic}

\iccvfinalcopy 

\ificcvfinal\fi

\begin{document}

\title{Pix2Pose: Pixel-Wise Coordinate Regression of Objects for 6D Pose Estimation}

\author{Kiru Park, Timothy Patten and Markus Vincze
\\
Vision for Robotics Laboratory, Automation and Control Institute, TU Wien, Austria\\
{\tt\small \{park, patten, vincze\}@acin.tuwien.ac.at}
}

\maketitle

\ificcvfinal\thispagestyle{empty}\fi

\makeatletter
\renewcommand{\paragraph}{%
  \@startsection{paragraph}{4}%
  {\z@}{1.0ex \@plus 0.5ex \@minus .5ex}{-1em}%
  {\normalfont\normalsize\bfseries}%
}
\newcommand\blfootnote[1]{%
  \begingroup
  \renewcommand\thefootnote{}\footnote{#1}%
  \addtocounter{footnote}{-1}%
  \endgroup
}

\makeatother

\begin{abstract}
Estimating the 6D pose of objects using only RGB images remains challenging because of problems such as occlusion and symmetries. It is also difficult to construct 3D models with precise texture without expert knowledge or specialized scanning devices. To address these problems, we propose a novel pose estimation method, Pix2Pose, that predicts the 3D coordinates of each object pixel without textured models. An auto-encoder architecture is designed to estimate the 3D coordinates and expected errors per pixel. These pixel-wise predictions are then used in multiple stages to form 2D-3D correspondences to directly compute poses with the PnP algorithm with RANSAC iterations. Our method is robust to occlusion by leveraging recent achievements in generative adversarial training to precisely recover occluded parts. Furthermore, a novel loss function, the transformer loss, is proposed to handle symmetric objects by guiding predictions to the closest symmetric pose. Evaluations on three different benchmark datasets containing symmetric and occluded objects show our method outperforms the state of the art using only RGB images.
\end{abstract}

\section{Introduction} \label{introduction}
Pose estimation of objects is an important task to understand the given scene and operate objects properly in robotic or augmented reality applications. The inclusion of depth images has induced significant improvements by providing precise 3D pixel coordinates~\cite{Hodan_2018_ECCV_bop,vidal20186d}. However, depth images are not always easily available, e.g., mobile phones and tablets, typical for augmented reality applications, offer no depth data. As such, substantial research is dedicated to estimating poses of known objects using RGB images only. \blfootnote{\noindent
\textcopyright 2019 IEEE. Personal use of this material is permitted. Permission from IEEE must be obtained for all other uses, in any current or future media, including reprinting/republishing this material for advertising or promotional purposes, creating new collective works, for resale or redistribution to servers or lists, or reuse of any copyrighted component of this work in other works. } 
\begin{figure}
\begin{center}
   \def\svgwidth{\linewidth}
   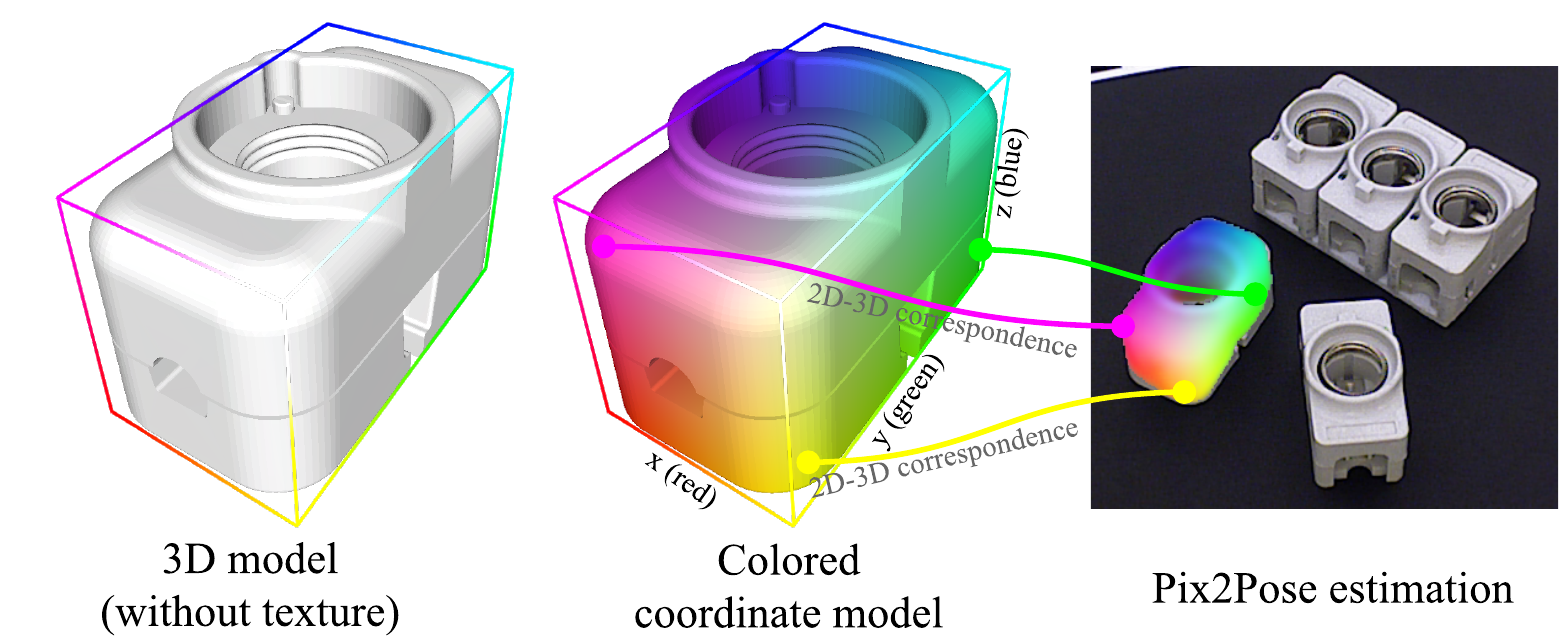
\end{center}
   \caption{An example of converting a 3D model to a colored coordinate model. Normalized coordinates of each vertex are directly mapped to red, green and blue values in the color space. Pix2Pose predicts these colored images to build a 2D-3D correspondence per pixel directly without any feature matching operation.}
\label{fig:3dcoord}
\end{figure}

A large body of work relies on the textured 3D model of an object, which is made by a 3D scanning device, e.g., BigBIRD Object Scanning Rig~\cite{alli2015YCB}, and provided by a dataset to render synthetic images for training~\cite{kehl2017ssd, Sundermeyer_2018_ECCV_implicit} or refinement~\cite{Li_2018_DeepIM, Manhardt_2018_ECCV_refinementTUM}. Thus, the quality of texture in the 3D model should be sufficient to render visually correct images. Unfortunately, this is not applicable to domains that do not have textured 3D models such as industry that commonly use texture-less CAD models. Since the texture quality of a reconstructed 3D model varies with method, camera, and camera trajectory during the reconstruction process, it is difficult to guarantee sufficient quality for training. Therefore, it is beneficial to predict poses without textures on 3D models to achieve more robust estimation regardless of the texture quality. 

Even though recent studies have shown great potential to estimate pose without textured 3D models using Convolutional Neural Networks (CNN)~\cite{cnn_pose:brachmann2016uncertainty_only_rgb,Do2018LieNetRM,rad2017bb8,Tekin_2018_CVPR}, a significant challenge is to estimate correct poses when objects are occluded or symmetric. Training CNNs is often distracted by symmetric poses that have similar appearance inducing very large errors in a na\"ive loss function. In previous work, a strategy to deal with symmetric objects is to limit the range of poses while rendering images for training~\cite{kehl2017ssd,Oberweger_2018_ECCV_heatmap} or simply to apply a transformation from the pose outside of the limited range to a symmetric pose within the range~\cite{rad2017bb8} for real images with pose annotations. This approach is sufficient for objects that have infinite and continuous symmetric poses on a single axis, such as cylinders, by simply ignoring the rotation about the axis. However, as pointed in \cite{rad2017bb8}, when an object has a finite number of symmetric poses, it is difficult to determine poses around the boundaries of view limits. For example, if a box has an angle of symmetry, $\pi$, with respect to an axis and a view limit between $0$ and $\pi$, the pose at $\pi+\alpha (\alpha \approx 0, \alpha > 0)$ has to be transformed to a symmetric pose at ${\alpha}$ even if the detailed appearance is closer to a pose at $\pi$. Thus, a loss function has to be investigated to guide pose estimations to the closest symmetric pose instead of explicitly defined view ranges. 

This paper proposes a novel method, \textit{Pix2Pose}, that can supplement any 2D detection pipeline for additional pose estimation. Pix2Pose predicts pixel-wise 3D coordinates of an object using RGB images without textured 3D models for training. The 3D coordinates of occluded pixels are implicitly estimated by the network in order to be robust to occlusion. A specialized loss function, the \textit{transformer loss}, is proposed to robustly train the network with symmetric objects. As a result of the prediction, each pixel forms a 2D-3D correspondence that is used to compute poses by the Perspective-n-Point algorithm~(PnP)~\cite{Lepetit2008epnp}.

 To summarize, the contributions of the paper are: 
\textbf{(1)} A novel framework for 6D pose estimation, Pix2Pose, that robustly regresses pixel-wise 3D coordinates of objects from RGB images using 3D models without textures during training. \textbf{(2)} A novel loss function, the transformer loss, for handling symmetric objects that have a finite number of ambiguous views. \textbf{(3)} Experimental results on three different datasets, LineMOD~\cite{linemode_hinterstoisser2012}, LineMOD Occlusion~\cite{brachmann2014learning_occlusion}, and T-Less~\cite{rgbddataset:tless}, showing that Pix2Pose outperforms the state-of-the-art methods even if objects are occluded or symmetric.

The remainder of this paper is organized as follows. A brief summary of related work is provided in Sec.~\ref{relatedworks}. Details of Pix2Pose and the pose prediction process are explained in Sec.~\ref{method} and Sec.~\ref{poseprediction}. Experimental results are reported in Sec.~\ref{evaluation} to compare our approach with the state-of-the-art methods. The paper concludes in Sec.~\ref{conclusion}.

\section{Related work} \label{relatedworks}
This section gives a brief summary of previous work related to pose estimation using RGB images. Three different approaches for pose estimation using CNNs are discussed and the recent advances of generative models are reviewed.

\paragraph{CNN based pose estimation} The first, and simplest, method to estimate the pose of an object using a CNN is to predict a representation of a pose directly such as the locations of projected points of 3D bounding boxes~\cite{rad2017bb8,Tekin_2018_CVPR}, classified view points~\cite{kehl2017ssd}, unit quaternions and translations~\cite{xiang2017posecnn}, or the Lie algebra representation, $so(3)$, with the translation of $z$-axis~\cite{Do2018LieNetRM}. Except for methods that predict projected points of the 3D bounding box, which requires further computations for the PnP algorithm, the direct regression is computationally efficient since it does not require additional computation for the pose. The drawback of these methods, however, is the lack of correspondences that can be useful to generate multiple pose hypotheses for the robust estimation of occluded objects. Furthermore, symmetric objects are usually handled by limiting the range of viewpoints, which sometimes requires additional treatments, e.g., training a CNN for classifying view ranges~\cite{rad2017bb8}. Xiang~\etal~\cite{xiang2017posecnn} propose a loss function that computes the average distance to the nearest points of transformed models in an estimated pose and an annotated pose. However, searching for the nearest 3D points is time consuming and makes the training process inefficient.

The second method is to match features to find the nearest pose template and use the pose information of the template as an initial guess~\cite{linemode_hinterstoisser2012}. Recently, Sundermeyer~\etal~\cite{Sundermeyer_2018_ECCV_implicit} propose an auto-encoder network to train implicit representations of poses without supervision using RGB images only. Manual handling of symmetric objects is not necessary for this work since the implicit representation can be close to any symmetric view. However, it is difficult to specify 3D translations using rendered templates that only give a good estimation of rotations. The size of the 2D bounding box is used to compute the $z$-component of 3D translation, which is too sensitive to small errors of 2D bounding boxes that are given from a 2D detection method.

The last method is to predict 3D locations of pixels or local shapes in the object space~\cite{cnn_pose:brachmann2016uncertainty_only_rgb,kehl2016deep,Oberweger_2018_ECCV_heatmap}. Brachmann~\etal~\cite{cnn_pose:brachmann2016uncertainty_only_rgb} regress 3D coordinates and predict a class for each pixel using the auto-context random forest. Oberwerger~\etal~\cite{Oberweger_2018_ECCV_heatmap} predict multiple heat-maps to localize the 2D projections of 3D points of objects using local patches. These methods are robust to occlusion because they focus on local information only. However, additional computation is required to derive the best result among pose hypotheses, which makes these methods slow. 

The method proposed in this paper belongs to the last category that predicts 3D locations of pixels in the object frame as in ~\cite{brachmann2014learning_occlusion,cnn_pose:brachmann2016uncertainty_only_rgb}. Instead of detecting an object using local patches from sliding windows, an independent 2D detection network is employed to provide areas of interest for target objects as performed in~\cite{Sundermeyer_2018_ECCV_implicit}.

\begin{figure*}
\begin{center}
   \def\svgwidth{\linewidth}
   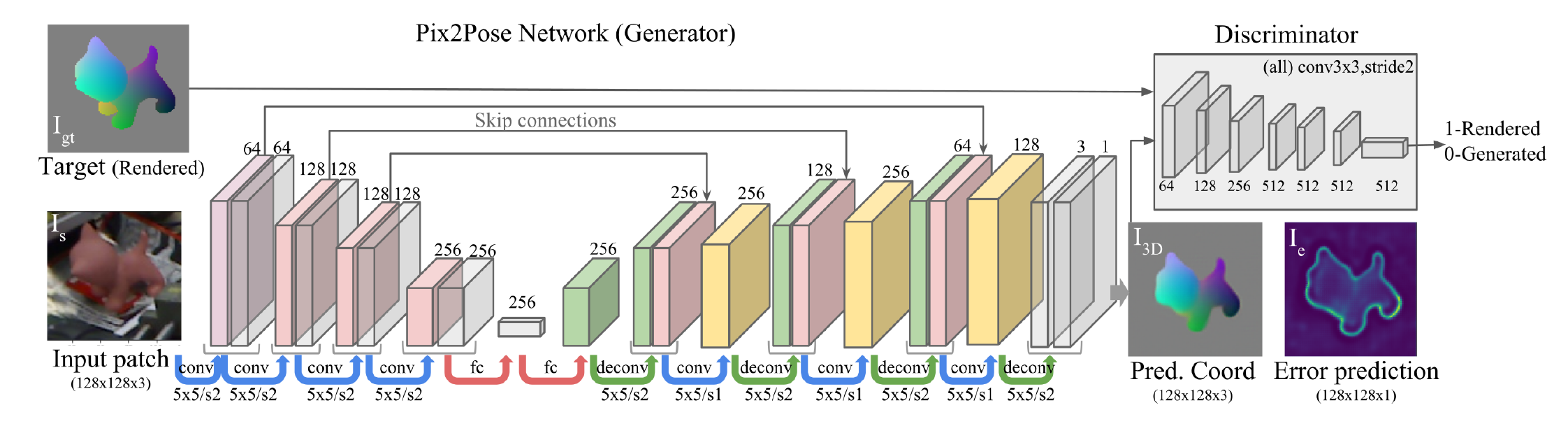
\end{center}
   \vspace{-13pt}
   \caption{An overview of the architecture of Pix2Pose and the training pipeline.}
\label{fig:network_arch}
\end{figure*}

\paragraph{Generative models} 
Generative models using auto-encoders have been used to de-noise~\cite{vincent2010stacked_denoising} or recover the missing parts of images~\cite{NIPS2012_4686_denoising_inpainting}. Recently, using Generative Adversarial Network~(GAN)~\cite{goodfellow2014generative} improves the quality of generated images that are less blurry and more realistic, which are used for the image-to-image translation ~\cite{Isola_2017_CVPR_pix2pix}, image in-painting and de-noising~\cite{iizuka2017globally_inpaintGAN,pathak2016context_inpaintGAN} tasks. Zakharov~\etal~\cite{zakharov2018keepit} propose a GAN based framework to convert a real depth image to a synthetic depth image without noise and background for classification and pose estimation. 

Inspired by previous work, we train an auto-encoder architecture with GAN to convert color images to coordinate values accurately as in the image-to-image translation task while recovering values of occluded parts as in the image in-painting task.

\section{Pix2Pose} \label{method}
This section provides a detailed description of the network architecture of Pix2Pose and loss functions for training. As shown in Fig.~\ref{fig:network_arch}, Pix2Pose predicts 3D coordinates of individual pixels using a cropped region containing an object. The robust estimation is established by recovering 3D coordinates of occluded parts and using all pixels of an object for pose prediction. A single network is trained and used for each object class. The texture of a 3D model is not necessary for training and inference.

\subsection{Network Architecture}
The architecture of the Pix2Pose network is described in Fig.~\ref{fig:network_arch}.
The input of the network is a cropped image $I_\textrm{s}$ using a bounding box of a detected object class. The outputs of the network are normalized 3D coordinates of each pixel $I_\textrm{3D}$ in the object coordinate and estimated errors $I_\textrm{e}$ of each prediction, $I_\textrm{3D},I_\textrm{e} = G(I_\textrm{s})$, where $G$ denotes the Pix2Pose network. The target output includes coordinate predictions of occluded parts, which makes the prediction more robust to partial occlusion. Since a coordinate consists of three values similar to RGB values in an image, the output $I_\textrm{3D}$ can be regarded as a color image. Therefore, the ground truth output is easily derived by rendering the colored coordinate model in the ground truth pose. An example of 3D coordinate values in a color image is visualized in Fig.~\ref{fig:3dcoord}. The error prediction $I_{e}$ is regarded as a confidence score of each pixel, which is directly used to determine outlier and inlier pixels before the pose computation.

 The cropped image patch is resized to 128$\times$128$px$ with three channels for RGB values. The sizes of filters and channels in the first four convolutional layers, the encoder, are the same as in~\cite{Sundermeyer_2018_ECCV_implicit}. To maintain details of low-level feature maps, skip connections~\cite{ronneberger2015u_unet} are added by copying the half channels of outputs from the first three layers to the corresponding symmetric layers in the decoder, which results in more precise estimation of pixels around geometrical boundaries. The filter size of every convolution and deconvolution layer is fixed to 5$\times$5 with stride 1 or 2 denoted as \textit{s1} or \textit{s2} in Fig.~\ref{fig:network_arch}. Two fully connected layers are applied for the bottle neck with 256 dimensions between the encoder and the decoder. The batch normalization~\cite{ioffe2015batchnorm} and the \textit{LeakyReLU} activation are applied to every output of the intermediate layers except the last layer. In the last layer, an output with three channels and the $\tanh$ activation produces a 3D coordinate image $I_\textrm{3D}$, and another output with one channel and the $\operatorname{sigmoid}$ activation estimates the expected errors $I_\textrm{e}$.

\begin{figure*}
\begin{center}
   \def\svgwidth{\linewidth}
   \includegraphics[width=1.0\textwidth]{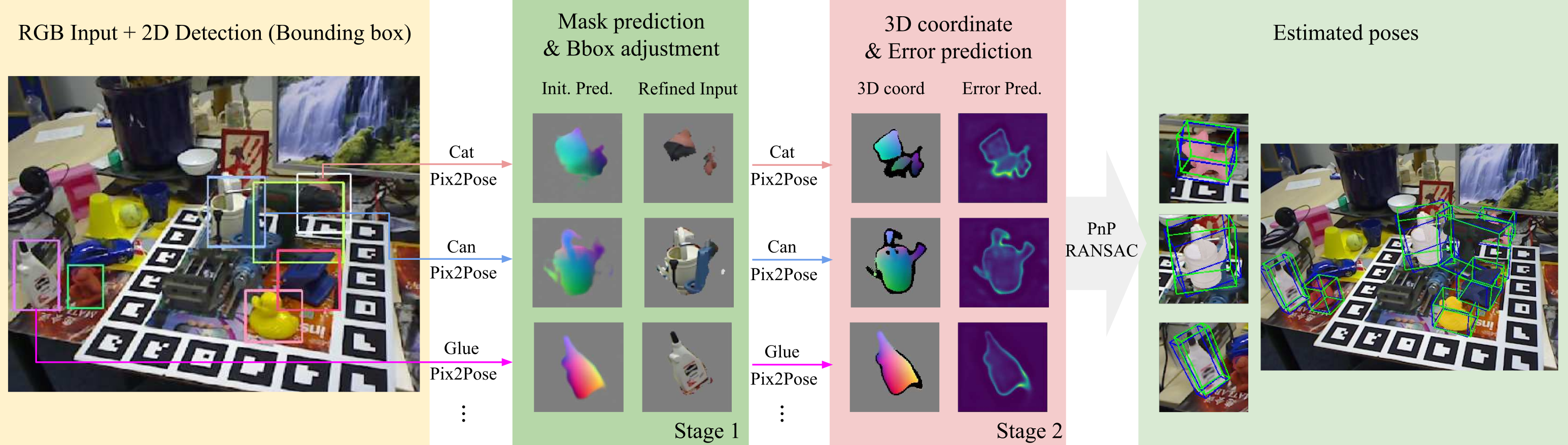} 
\end{center}
\vspace{-5pt}
   \caption{An example of the pose estimation process. An image and 2D detection results are the input. In the first stage, the predicted results are used to specify important pixels and adjust bounding boxes while removing backgrounds and uncertain pixels. In the second stage, pixels with valid coordinate values and small error predictions are used to estimate poses using the PnP algorithm with RANSAC. Green and blue lines in the result represent 3D bounding boxes of objects in ground truth poses and estimated poses.
   }
\label{fig:detection_process}
\end{figure*}
\subsection{Network Training}

The main objective of training is to predict an output that minimizes errors between a target coordinate image and a predicted image while estimating expected errors of each pixel.

\paragraph{Transformer loss for 3D coordinate regression} \label{transformer}
To reconstruct the desired target image, the average L1 distance of each pixel is used. Since pixels belonging to an object are more important than the background, the errors under the object mask are multiplied by a factor of $\beta~(\geq1)$ to weight errors in the object mask. The basic reconstruction loss $\mathcal{L}_\textrm{r}$ is defined as,
\begin{equation} \label{eq:recont_loss}    
\mathcal{L}_\textrm{r} = \frac{1}{n} \Big[ \beta\sum_{i \in M}||I_\textrm{3D}^i-I_\textrm{gt}^i||_1 + \sum_{i \notin M} ||I_\textrm{3D}^i-I_\textrm{gt}^i||_1 \Big],
\end{equation} 
where $n$ is the number of pixels, $I_\textrm{gt}^i$ is the $i^\textrm{th}$ pixel of the target image, and $M$ denotes an object mask of the target image, which includes pixels belonging to the object when it is fully visible. Therefore, this mask also contains the occluded parts to predict the values of invisible parts for robust estimation of occluded objects.

The loss above cannot handle symmetric objects since it penalizes pixels that have larger distances in the 3D space without any knowledge of the symmetry. Having the advantage of predicting pixel-wise coordinates, the 3D coordinate of each pixel is easily transformed to a symmetric pose by multiplying a 3D transformation matrix to the target image directly. Hence, the loss can be calculated for a pose that has the smallest error among symmetric pose candidates as formulated by, 
\begin{equation} \label{eq:sym_loss}    
\mathcal{L}_\textrm{3D}= \min_{p \in \textrm{sym}}  \mathcal{L}_\textrm{r}(I_\textrm{3D},R_{p}I_{gt}), 
\end{equation} 
where $R_{p} \in \mathbb{R}^{3\textrm{x}3}$ is a transformation from a pose to a symmetric pose in a pool of symmetric poses, $sym$, including an identity matrix for the given pose. The pool $sym$ is assumed to be defined before the training of an object. This novel loss, the transformer loss, is applicable to any symmetric object that has a finite number of symmetric poses. This loss adds only a tiny effort for computation since a small number of matrix multiplications is required. The transformer loss in Eq.~\ref{eq:sym_loss} is applied instead of the basic reconstruction loss in Eq.~\ref{eq:recont_loss}. The benefit of the transformer loss is analyzed in Sec.~\ref{ablation:transform}.

\paragraph{Loss for error prediction}
The error prediction $I_e$ estimates the difference between the predicted image $I_\textrm{3D}$ and the target image $I_\textrm{gt}$. This is identical to the reconstruction loss $\mathcal{L}_\textrm{r}$ with $\beta=1$ such that pixels under the object mask are not penalized. Thus, the error prediction loss $\mathcal{L}_\textrm{e}$ is written as, 
\begin{equation} \label{eq:error_loss}    
\mathcal{L}_\textrm{e} =\frac{1}{n} \sum_{i}||I_\textrm{e}^{i} - \textrm{min}\big[\mathcal{L}_\textrm{r}^{i},1\big]||^2_2, \beta = 1.
\end{equation} 
The error is bounded to the maximum value of the $\operatorname{sigmoid}$ function.

\paragraph{Traininig with GAN} As discussed in Sec.~\ref{relatedworks}, the network training with GAN generates more precise and realistic images in a target domain using images of another domain~\cite{Isola_2017_CVPR_pix2pix}. The task for Pix2Pose is similar to this task since it converts a color image to a 3D coordinate image of an object. Therefore, the discriminator and the loss function of GAN~\cite{goodfellow2014generative}, $\mathcal{L}_{\textrm{GAN}}$, is employed to train the network. As shown in Fig.~\ref{fig:network_arch}, the discriminator network attempts to distinguish whether the 3D coordinate image is rendered by a 3D model or is estimated. The loss is defined as, 
\begin{equation} \label{eq:gan_loss}    
\mathcal{L}_{\textrm{GAN}} = \log D(I_{gt}) + \log (1-D(G(I_\textrm{src}))),
\end{equation}
where D denotes the discriminator network. Finally, the objective of the training with GAN is formulated as, 
\begin{equation} \label{eq:loss_all}    
G^* = \textrm{arg}\min_{G} \max_{D}  \mathcal{L}_\textrm{GAN} (G,D) + \lambda_1 \mathcal{L}_\textrm{3D} (G) + \lambda_2 \mathcal{L}_\textrm{e}(G),
\end{equation} 
where $\lambda_1$ and $\lambda_2$ denote weights to balance different tasks. 

\section{Pose prediction} \label{poseprediction}
This section gives a description of the process that computes a pose using the output of the Pix2Pose network. The overview of the process is shown in Fig.~\ref{fig:detection_process}.
Before the estimation, the center, width, and height of each bounding box are used to crop the region of interest and resize it to the input size, 128$\times$128$px$. The width and height of the region are set to the same size to keep the aspect ratio by taking the larger value. Then, they are multiplied by a factor of 1.5 so that the cropped region potentially includes occluded parts. The pose prediction is performed in two stages and the identical network is used in both stages. The first stage aligns the input bounding box to the center of the object which could be shifted due to different 2D detection methods. It also removes unnecessary pixels (background and uncertain) that are not preferred by the network. The second stage predicts a final estimation using the refined input from the first stage and computes the final pose.

\paragraph{Stage 1: Mask prediction and Bbox Adjustment}
In this stage, the predicted coordinate image $I_\textrm{3D}$ is used for specifying pixels that belong to the object including the occluded parts by taking pixels with non-zero values. The error prediction is used to remove the uncertain pixels if an error for a pixel is larger than the outlier threshold $\theta_{o}$. The valid object mask is computed by taking the union of pixels that have non-zero values and pixels that have lower errors than $\theta_{o}$. The new center of the bounding box is determined with the centroid of the valid mask. As a result, the output of the first stage is a refined input that only contains pixels in the valid mask cropped from a new bounding box. Examples of outputs of the first stage are shown in Fig.~\ref{fig:detection_process}. The refined input possibly contains the occluded parts when the error prediction is below the outlier threshold $\theta_{o}$, which means the coordinates of these pixels are easy to predict despite occlusions.

\paragraph{Stage 2: Pixel-wise 3D coordinate regression with errors}
The second estimation with the network is performed to predict a coordinate image and expected error values using the refined input as depicted in Fig.~\ref{fig:detection_process}. Black pixels in the 3D coordinate samples denote points that are removed when the error prediction is larger than the inlier threshold $\theta_{i}$ even though points have non-zero coordinate values. In other words, pixels that have non-zero coordinate values with smaller error predictions than $\theta_{i}$ are used to build 2D-3D correspondences. Since each pixel already has a value for a 3D point in the object coordinate, the 2D image coordinates and predicted 3D coordinates directly form correspondences. Then, applying the PnP algorithm~\cite{Lepetit2008epnp} with RANdom SAmple Consensus (RANSAC)~\cite{Ransac} iteration computes the final pose by maximizing the number of inliers that have lower re-projection errors than a threshold $\theta_{re}$. It is worth mentioning that there is no rendering involved during the pose estimation since Pix2Pose does not assume textured 3D models. This also makes the estimation process fast. 

\section{Evaluation} \label{evaluation}
In this section, experiments on three different datasets are performed to compare the performance of Pix2Pose to state-of-the-art methods. The evaluation using LineMOD~\cite{linemode_hinterstoisser2012} shows the performance for objects without occlusion in the single object scenario. For the multiple object scenario with occlusions, LineMOD Occlusion~\cite{brachmann2014learning_occlusion} and T-Less ~\cite{rgbddataset:tless} are used. The evaluation on T-Less shows the most significant benefit of Pix2Pose since T-Less provides texture-less CAD models and most of the objects are symmetric, which is more challenging and common in industrial domains.

\begin{figure}
\begin{center}
   \def\svgwidth{\linewidth}
   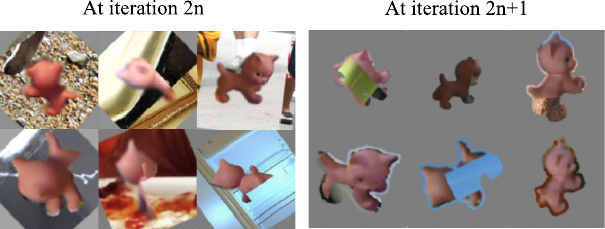
\end{center}
    \vspace{-5pt}
   \caption{Examples of mini-batches for training. A mini-batch is altered for every training iteration. Left: images for the first stage, Right: images for the second stage.}
 
\label{fig:batchsample}
\end{figure}

\subsection{Augmentation of training data} 
A small number of real images are used for training with various augmentations. Image pixels of objects are extracted from real images and pasted to background images that are randomly picked from the Coco dataset~\cite{lin2014mscoco}. After applying the color augmentations on the image, the borderlines between the object and the background are blurred to make smooth boundaries. A part of the object area is replaced by the background image to simulate occlusion. Lastly, a random rotation is applied to both the augmented color image and the target coordinate image. The same augmentation is applied to all evaluations except sizes of occluded areas that need to be larger for datasets with occlusions, LineMOD Occlusion and T-Less. Sample augmentated images are shown in Fig.~\ref{fig:batchsample}. As explained in Sec.~\ref{poseprediction}, the network recognizes two types of inputs, with background in the first stage and without background pixels in the second stage. Thus, a mini-batch is altered for every iteration as shown in Fig.~\ref{fig:batchsample}. Target coordinate images are rendered before training by placing the object in the ground truth poses using the colored coordinate model as in Fig.~\ref{fig:3dcoord}. 

\setlength{\tabcolsep}{4.5pt}
\begin{table*}[ht]
\begin{center}
\begin{tabular}{c | c c c c c c c c c c c c c|c}
 \space &ape & bvise & cam&can & cat & driller & duck & e.box* & glue* & holep & iron & lamp & phone & avg\\
 \hline
 \textbf{Pix2Pose} & \textbf{58.1} & \textbf{91.0} & \textbf{60.9} & \textbf{84.4} & \textbf{65.0} & \textbf{76.3} & \textbf{43.8} & \textbf{96.8} & 79.4 & \textbf{74.8} & \textbf{83.4} & \textbf{82.0} & 45.0 & \textbf{72.4}\\

Tekin~\cite{Tekin_2018_CVPR} & 21.6 & 81.8 & 36.6 & 68.8 & 41.8 & 63.5 & 27.2 & 69.6 &\textbf{80.0} & 42.6 &75.0 & 71.1 & \textbf{47.7} & 56.0\\
Brachmann~\cite{cnn_pose:brachmann2016uncertainty_only_rgb} & 33.2 & 64.8 & 38.4 & 62.9& 42.7& 61.9&30.2 & 49.9 & 31.2 & 52.8 & 80.0 & 67.0 & 38.1 & 50.2 \\
BB8~\cite{rad2017bb8} & 27.9 & 62.0 & 40.1 & 48.1 &45.2 & 58.6 & 32.8 & 40.0 & 27.0 & 42.4 & 67.0 & 39.9 & 35.2 & 43.6 \\
\hline
Lienet$^{30\%}$~\cite{Do2018LieNetRM} & 38.8 & 71.2 & 52.5 & 86.1 &66.2 & 82.3 & 32.5 & 79.4 & 63.7 & 56.4 & 65.1 & 89.4 & 65.0 & 65.2 \\
BB8$^{\textrm{ref}}$ ~\cite{rad2017bb8} & 40.4 & 91.8 & 55.7 & 64.1 &62.6 & 74.4 & 44.3 & 57.8 & 41.2 & 67.2 & 84.7 & 76.5 & 54.0  & 62.7\\
Implicit$^{\textrm{syn}}$~\cite{Sundermeyer_2018_ECCV_implicit} & 4.0 & 20.9 & 30.5 & 35.9 &17.9 & 24.0 & 4.9 & 81.0 & 45.5 & 17.6 & 32.0 & 60.5 & 33.8 & 31.4\\
SSD-6D$^\textrm{syn/ref}$~\cite{kehl2017ssd} & 65 & 80 & 78 & 86 & 70 & 73 &66 & 100 & 100 & 49 & 78 & 73 & 79 & 76.7\\
Rad$^\textrm{syn/ref}$~\cite{Rad_2018_CVPR} & - & - & - & - & - & - &- & - & - & - & - & - & - & \textit{\textbf{78.7}}\\

\end{tabular}
\end{center}
\caption{LineMOD: Percentages of correctly estimated poses (AD\{D$\vert$I\}-10\%). ($^{30\%}$) means the training images are obtained from 30\% of test sequences that are two times larger than ours. ($^{\textrm{ref}}$) denotes the results are derived after iterative refinement using textured 3D models for rendering. ($^\textrm{syn}$) indicates the method uses synthetically rendered images for training that also needs textured 3D models.}
\label{table_linemod}
\end{table*}

\subsection{Implementation details}
For training, the batch size of each iteration is set to 50, the Adam optimizer~\cite{kingma2014adam} is used with initial learning rate of 0.0001 for 25K iterations. The learning rate is multiplied by a factor of 0.1 for every 12K iterations. Weights of loss functions in Eq.~\ref{eq:recont_loss} and Eq.~\ref{eq:loss_all} are: $\beta$=3, $\lambda_1$=100 and $\lambda_2$=50.
For evaluation, a 2D detection network and Pix2Pose networks of all object candidates in test sequences are loaded to the GPU memory, which requires approximately 2.2GB for the LineMOD Occlusion experiment with eight objects. The standard parameters for the inference are: $\theta_i$=0.1, $\theta_{o}$=[0.1, 0.2, 0.3], and $\theta_{re}$=3. Since the values of error predictions are biased by the level of occlusion in the online augmentation and the shape and size of each object, the outlier threshold $\theta_o$ in the first stage is determined among three values to include more numbers of visible pixels while excluding noisy pixels using samples of training images with artificial occlusions. More details about parameters are given in the supplementary material. The training and evaluations are performed with an Nvidia GTX 1080 GPU and i7-6700K CPU.   

\paragraph{2D detection network}
An improved Faster R-CNN~\cite{he2017mask,cnn:fastercnn} with Resnet-101~\cite{he2016deep_resnet} and Retinanet~\cite{Lin_2017_ICCV_retinanet} with Resnet-50 are employed to provide classes of detected objects with 2D bounding boxes for all target objects of each evaluation. The networks are initialized with pre-trained weights using the Coco dataset~\cite{lin2014mscoco}.  The same set of real training images is used to generate training images. Cropped patches of objects in real images are pasted to random background images to generate training images that contain multiple classes in each image.

\subsection{Metrics}
A standard metric for LineMOD, AD\{D$\vert$I\}, is mainly used for the evaluation~\cite{linemode_hinterstoisser2012}. This measures the average distance of vertices between a ground truth pose and an estimated pose. For symmetric objects, the average distance to the nearest vertices is used instead. The pose is considered correct when the error is less than 10\% of the maximum 3D diameter of an object.

For T-Less, the Visible Surface Discrepancy (VSD) is used as a metric since the metric is employed to benchmark various 6D pose estimation methods in~\cite{Hodan_2018_ECCV_bop}. This metric measures distance errors of visible parts only, which makes the metric invariant to ambiguities caused by symmetries and occlusion. As in previous work, the pose is regarded as correct when the error is less than 0.3 with $\tau$=20\textit{mm} and $\delta$=15\textit{mm}.

\subsection{LineMOD} \label{eval_linemod}
For training, test sequences are separated into a training and test set. The divided set of each sequence is identical to the work of~\cite{cnn_pose:brachmann2016uncertainty_only_rgb, Tekin_2018_CVPR}, which uses 15\% of test scenes, approximately less than 200 images per object, for training. A detection result, using Faster R-CNN, of an object with the highest score in each scene is used for pose estimation since the detection network produces multiple results for all 13 objects. For the symmetric objects, marked with (*) in Table~\ref{table_linemod}, the pool of symmetric poses $sym$ is defined as, $sym$= $[I,R^{\pi}_z]$, where $R^{\pi}_{z}$ represents a transformation matrix of rotation with $\pi$ about the $z$-axis.

The upper part of Table~\ref{table_linemod} shows Pix2Pose significantly outperforms state-of-the-art methods that use the same amount of real training images without textured 3D models. Even though methods on the bottom of Table~\ref{table_linemod} use a larger portion of training images, use textured 3D models for training or pose refinement, our method shows competitive results against these methods. The results on symmetric objects show the best performance among methods that do not perform pose refinement. This verifies the benefit of the transformer loss, which improves the robustness of initial pose predictions for symmetric objects.

\setlength{\tabcolsep}{4pt}
\begin{table} 
\begin{center}
\begin{tabular}{c | c| c|c|c }
 \begin{tabular}{@{}c@{}}Method\\\space\end{tabular} & 
 \begin{tabular}{@{}c@{}}\textbf{Pix2Pose}\\\space\end{tabular}
 &\begin{tabular}{@{}c@{}}Oberweger$^\dagger$ \\ \cite{Oberweger_2018_ECCV_heatmap}\end{tabular}
 & \begin{tabular}{@{}c@{}}PoseCNN$^\dagger$\\ \cite{xiang2017posecnn}\end{tabular} &
 \begin{tabular}{@{}c@{}}Tekin\\ \cite{Tekin_2018_CVPR}\end{tabular} \\
 \hline
 ape& \textbf{22.0} & 17.6 & 9.6 & 2.48\\
 can& 44.7 & \textbf{53.9} & 45.2 & 17.48\\
 cat& \textbf{22.7} & 3.31 & 0.93 & 0.67\\
 driller& 44.7 & \textbf{62.4} & 41.4 &7.66\\
 duck& 15.0 & 19.2 & \textbf{19.6} & 1.14\\
 eggbox*& 25.2 & \textbf{25.9} &22.0 & -\\
 glue*& 32.4 & \textbf{39.6} & 38.5 & 10.08\\
 holep& \textbf{49.5} &21.3 & 22.1 & 5.45\\
 \hline
 Avg& \textbf{32.0} & 30.4 & 24.9 & 6.42\\
\end{tabular}
\end{center}
\caption{LineMOD Occlusion: object recall (AD\{D$\vert$I\}-10\%). ($^\dagger$) indicates the method uses synthetically rendered images and real images for training, which has better coverage of viewpoints.}
\label{table_occ}
\end{table}

\subsection{LineMOD Occlusion}
LineMOD Occlusion is created by annotating eight objects in a test sequence of LineMOD. Thus, the test sequences of eight objects in LineMOD are used for training without overlapping with test images. Faster R-CNN is used as a 2D detection pipeline.

As shown in Table~\ref{table_occ}, Pix2Pose significantly outperforms the method of~\cite{Tekin_2018_CVPR} using only real images for training. Furthermore, Pix2Pose outperforms the state of the art on three out of eight objects. On average it performs best even though methods of \cite{Oberweger_2018_ECCV_heatmap} and \cite{xiang2017posecnn} use more images that are synthetically rendered by using textured 3D models of objects. Although these methods cover more various poses than the given small number of images, Pix2Pose robustly estimates poses with less coverage of training poses.

\subsection{T-Less}
In this dataset, a CAD model without textures and a reconstructed 3D model with textures are given for each object. Even though previous work uses reconstructed models for training, to show the advantage of our method, CAD models are used for training (as shown in Fig.~\ref{fig:3dcoord}) with real training images provided by the dataset. To minimize the gap of object masks between a real image and a rendered scene using a CAD model, the object mask of the real image is used to remove pixels outside of the mask in the rendered coordinate images. The pool of symmetric poses $sym$ of objects is defined manually similar to the \textit{eggbox} in the LineMOD evaluation for box-like objects such as \textit{obj-05}. For cylindrical objects such as \textit{obj-01}, the rotation component of the $z$-axis is simply ignored and regarded as a non-symmetric object. The experiment is performed based on the protocol of~\cite{Hodan_2018_ECCV_bop}. Instead of a subset of the test sequences in~\cite{Hodan_2018_ECCV_bop}, full test images are used to compare with the state of the art~\cite{Sundermeyer_2018_ECCV_implicit}. Retinanet is used as a 2D detection method and objects visible more than 10\% are considered as estimation targets~\cite{Hodan_2018_ECCV_bop,Sundermeyer_2018_ECCV_implicit}. 

The result in Table~\ref{table_tless} shows Pix2Pose outperforms the-state-of-the-art method that uses RGB images only by a significant margin. The performance is also better than the best learning-based methods~\cite{cnn_pose:brachmann2016uncertainty_only_rgb, kehl2016deep} in the benchmark~\cite{Hodan_2018_ECCV_bop}. Although these methods use color and depth images to refine poses or to derive the best pose among multiple hypotheses, our method, that predicts a single pose per detected object, performs better than these methods without refinement using depth images. 

\begin{table} 
\begin{center}
\begin{tabular}{c | c| c |c | c}
  Input &  \multicolumn{2}{c|}{RGB only} & \multicolumn{2}{c}{RGB-D} \\
 \hline
 
 Method & \textbf{Pix2Pose} & \begin{tabular}{@{}c@{}}Implicit\\\cite{Sundermeyer_2018_ECCV_implicit}\end{tabular}
 
 & 
 \begin{tabular}{@{}c@{}}Kehl\\\cite{kehl2016deep}\end{tabular}&
 \begin{tabular}{@{}c@{}}Brachmann\\\cite{cnn_pose:brachmann2016uncertainty_only_rgb}\end{tabular}\\

 \hline
 Avg & \textbf{29.5} & 18.4&24.6&17.8\\

\end{tabular}
\end{center}
\caption{T-Less: object recall ($e_\textrm{VSD} < $ 0.3, $\tau$ = 20mm) on all test scenes using PrimeSense. Results of~\cite{kehl2016deep} and ~\cite{cnn_pose:brachmann2016uncertainty_only_rgb} are cited from \cite{Hodan_2018_ECCV_bop}. Object-wise results are included in the supplement material.}
\label{table_tless}
\end{table}

\subsection{Ablation studies} \label{ablation}
In this section, we present ablation studies by answering four important questions that clarify the contribution of each component in the proposed method.

\begin{figure}
\begin{center}
   \def\svgwidth{\linewidth}
   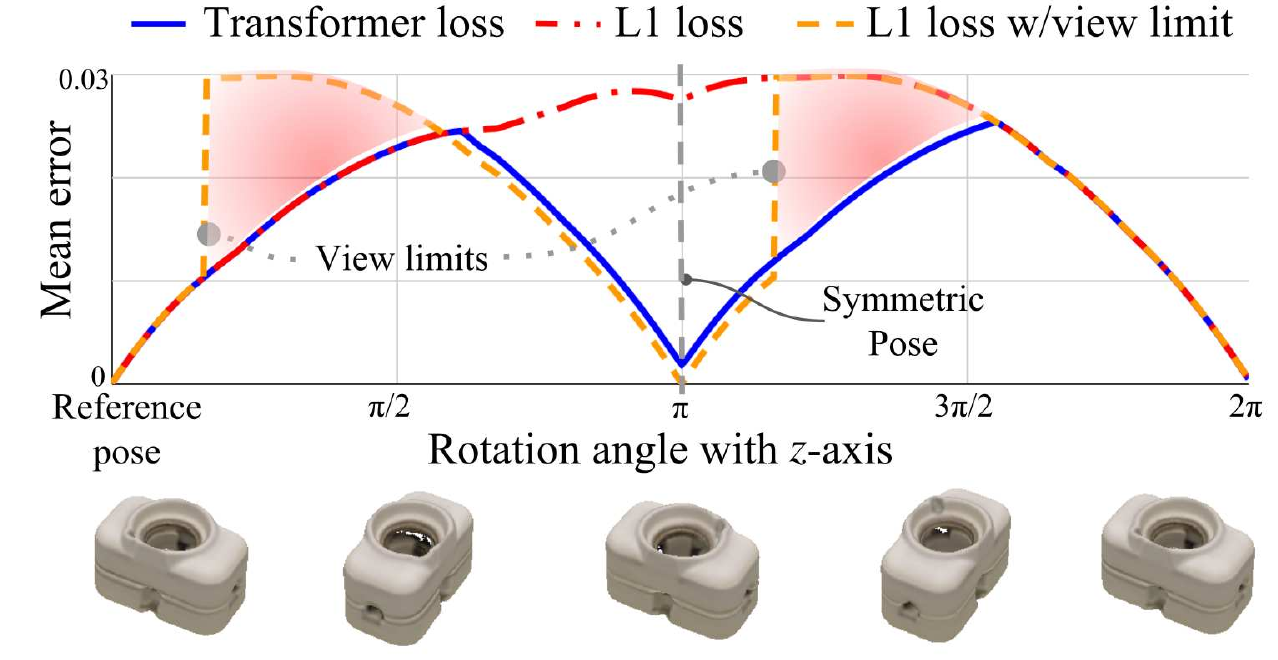
\end{center}
\vspace{-8pt}
   \caption{Variation of the reconstruction loss for a symmetric object with respect to $z$-axis rotation using obj-05 in T-Less~\cite{rgbddataset:tless}.}
\label{fig:transformer}
\end{figure}

\paragraph{How does the transformer loss perform?} \label{ablation:transform}
The \textit{obj-05} in T-Less is used to analyze the variation of loss values with respect to symmetric poses and to show the contribution of the transformer loss. To see the variation of loss values, 3D coordinate images are rendered while rotating the object around the $z$-axis. Loss values are computed using the coordinate image of a reference pose as a target output $I_\textrm{gt}$ and images of other poses as predicted outputs $I_\textrm{3D}$ in Eq.~\ref{eq:recont_loss} and Eq.~\ref{eq:sym_loss}. As shown in Fig.~\ref{fig:transformer}, the L1 loss in Eq.~\ref{eq:recont_loss} produces large errors for symmetric poses around $\pi$, which is the reason why the handling of symmetric objects is required. On the other hand, the value of the transformer loss produces minimum values on $0$ and $\pi$, which is expected for \textit{obj-05} with an angle of symmetry of $\pi$. The result denoted by \textit{view limits} shows the value of the L1 loss while limiting the $z$-component of rotations between $0$ and $\pi$. The pose that exceeds this limit is rotated to a symmetric pose. As discussed in Sec.~\ref{introduction}, values are significantly changed at the angles of view limits and over-penalize poses under areas with red in Fig.~\ref{fig:transformer}, which causes noisy predictions of poses around these angles. The results in Table~\ref{table_tless_obj05} show the transformer loss significantly improves the performance compared to the L1 loss with the view limiting strategy and the L1 loss without handling symmetries.

\newcolumntype{C}[1]{>{\centering\arraybackslash}p{#1}}
\begin{table} 
\begin{center}
\begin{tabular}{C{2.4cm}|C{2.4cm}| C{2.4cm} }
  \textrm{Transformer loss} &  L1-view limits & L1 \\
 \hline
 \textbf{55.2} &47.2&33.4\\
\end{tabular}
\end{center}
\vspace{-2pt}
\caption{Recall ($e_\textrm{vsd}<0.3$) of obj-05 in T-Less using different reconstruction losses for training.}
\label{table_tless_obj05}
\end{table}

\begin{figure}
\begin{center}
   \def\svgwidth{\linewidth}
   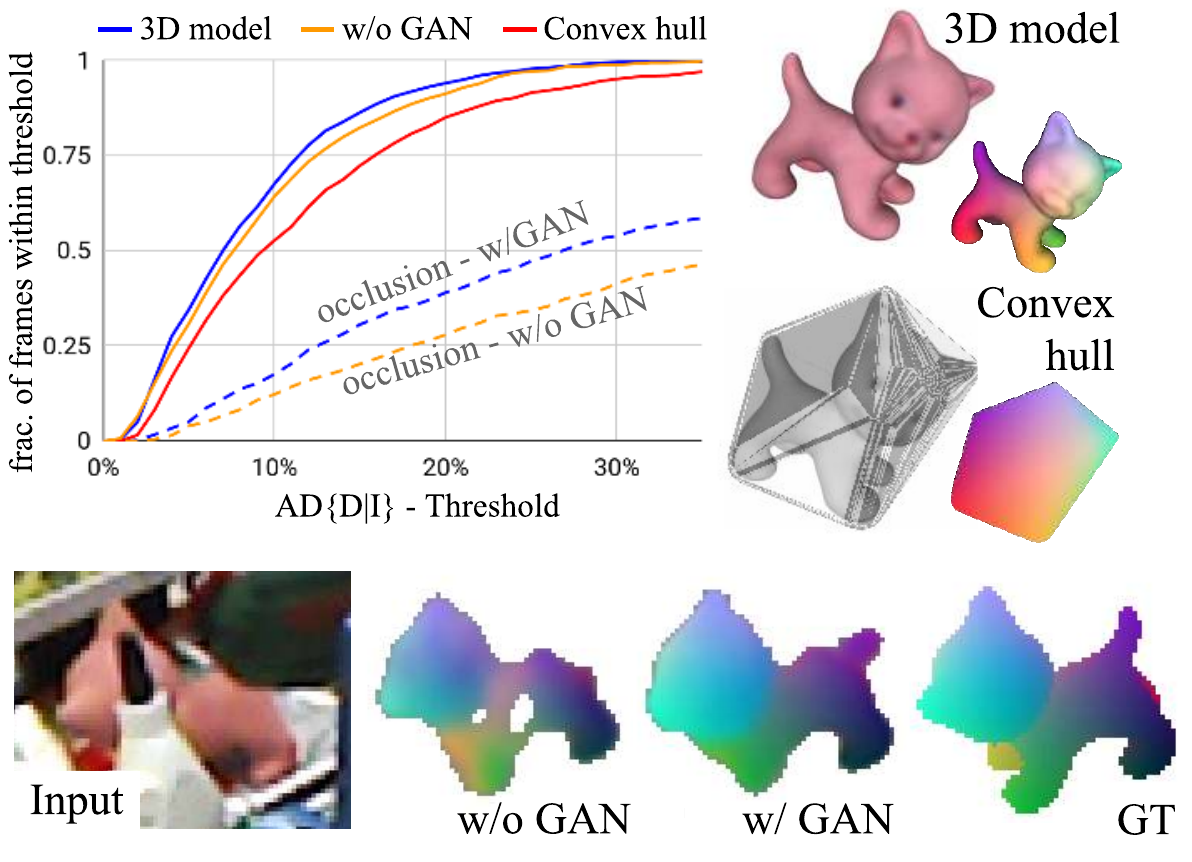
\end{center}
   \vspace{-7pt}
   \caption{Top: the fraction of frames within AD\{D$\vert$I\} thresholds for the \textit{cat} in LineMOD. The larger area under a curve means better performance. Bottom: qualitative results with/without GAN.}
\label{fig:convex_hull}
\end{figure}

\paragraph{What if the 3D model is not precise?}
The evaluation on T-Less already shows the robustness to 3D CAD models that have small geometric differences with real objects. However, it is often difficult to build a 3D model or a CAD model with refined meshes and precise geometries of a target object. Thus, a simpler 3D model, a convex hull covering out-bounds of the object, is used in this experiment as shown in Fig.~\ref{fig:convex_hull}. The training and evaluation are performed in the same way for the LineMOD evaluation with synchronization of object masks using annotated masks of real images. As shown in the top-left of Fig.~\ref{fig:convex_hull}, the performance slightly drops when using the convex hull. However, the performance is still competitive with methods that use 3D bounding boxes of objects, which means that Pix2Pose uses the details of 3D coordinates for robust estimation even though 3D models are roughly reconstructed.

\paragraph{Does GAN improve results?}
The network of Pix2Pose can be trained without GAN by removing the GAN loss in the final loss function in Eq.~\ref{eq:loss_all}. Thus, the network only attempts to reconstruct the target image without trying to trick the discriminator. To compare the performance, the same training procedure is performed without GAN until the loss value excluding the GAN loss reaches 
the same level. Results in the top-left in Fig.~\ref{fig:convex_hull} shows the fraction of correctly estimated poses with varied thresholds for the ADD metric. Solid lines show the performance on the original LineMOD test images, which contains fully visible objects, and dashed lines represent the performance on the same test images with artificial occlusions that are made by replacing 50\% of areas in each bounding box with zero. There is no significant change in the performance when objects are fully visible. However, the performance drops significantly without GAN when objects are occluded. Examples in the bottom of Fig.~\ref{fig:convex_hull} also show training with GAN produces robust predictions on occluded parts.

\paragraph{Is Pix2Pose robust to different 2D detection networks?}
Table~\ref{table_2d_detection} reports the results using different 2D detection networks on LineMOD. Retinanet and Faster R-CNN are trained using the same training images used in the LineMOD evaluation. In addition, the public code and trained weights of SSD-6D~\cite{kehl2017ssd} are used to derive 2D detection results while ignoring pose predictions of the network. It is obvious that pose estimation results are proportional to 2D detection performances. On the other hand, the portion of correct poses on good bounding boxes (those that overlap more than 50\% with ground truth) does not change significantly. This shows that Pix2Pose is robust to different 2D detection results when a bounding box overlaps the target object sufficiently. This robustness is accomplished by the refinement in the first stage that extracts useful pixels with a re-centered bounding box from a test image. Without the two stage approach, the performance significantly drops to 41\% on LineMOD when the output of the network in the first stage is used directly for the PnP computation. 
\setlength{\tabcolsep}{3pt}
\begin{table} 
\begin{center}
\begin{tabular}{c | c| c|c |c }
 \space &  \begin{tabular}{@{}c@{}}SSD-6D\\ \cite{kehl2017ssd}\end{tabular} &   \begin{tabular}{@{}c@{}}Retina~\\ \cite{Lin_2017_ICCV_retinanet}\end{tabular} &  
   \begin{tabular}{@{}c@{}}R-CNN~\\ \cite{cnn:fastercnn}\end{tabular} &
   \begin{tabular}{@{}c@{}}GT \\ bbox\end{tabular}\\
 \hline
 2D bbox  & 89.1 & 97.7&98.6& 100 \\
 \hline
 6D pose  & 64.0 & 71.1& 72.4& 74.7\\
 6D pose/2D bbox & 70.9 & 72.4 & 73.2& 74.7
\end{tabular}
\end{center}
\vspace{-2pt}
\caption{Average percentages of correct 2D bounding boxes (IoU$>$0.5) and correct 6D poses (ADD-10\%) on LineMOD using different 2D detection methods. The last row reports the percentage of correctly estimated poses on scenes that have correct bounding boxes (IoU$>$0.5).}
\label{table_2d_detection}
\end{table}
\subsection{Inference time} \label{inferencetime}
The inference time varies according to the 2D detection networks. Faster R-CNN takes 127ms and Retinanet takes 76ms to detect objects from an image with 640$\times$480$px$. The pose estimation for each bounding box takes approximately 25-45ms per region. Thus, our method is able to estimate poses at 8-10 fps with Retinanet and 6-7 fps with Faster R-CNN in the single object scenario.

\section{Conclusion} \label{conclusion}
This paper presented a novel architecture, Pix2Pose, for 6D object pose estimation from RGB images. Pix2Pose addresses several practical problems that arise during pose estimation: the difficulty of generating real-world 3D models with high-quality texture as well as robust pose estimation of occluded and symmetric objects. Evaluations with three challenging benchmark datasets show that Pix2Pose significantly outperforms state-of-the-art methods while solving these aforementioned problems.

Our results reveal that many failure cases are related to unseen poses that are not sufficiently covered by training images or the augmentation process. Therefore, future work will investigate strategies to improve data augmentation to more broadly cover pose variations using real images in order to improve estimation performance. Another avenue for future work is to generalize the approach to use a single network to estimate poses of various objects in a class that have similar geometry but different local shape or scale.

\small{\paragraph{Acknowledgment}
The research leading to these results has received funding from the Austrian Science Foundation (FWF) under grant agreement No. I3967-N30 (BURG) and No. I3969-N30 (InDex), and Aelous Robotics, Inc.}

{\small
\bibliographystyle{ieee_fullname}
\bibliography{Citation2019}
}

\onecolumn
\appendix
\section{Detail parameters}

\subsection{Data augmentation for training}

\begin{table}[hbt]
\begin{center}
\begin{tabular}{ c| c| c| c}
  \hline
  Add(each channel) & Contrast normalization &Multiply  & Gaussian Blur\\
 
 \hline
 $\mathcal{U}$(-15, 15) & $\mathcal{U}$(0.8, 1.3) & $\mathcal{U}$(0.8, 1.2)(per channel chance=0.3) & $\mathcal{U}$(0.0, 0.5) \\
  \hline
\end{tabular}
\end{center}
\caption{Color augmentation}
\end{table}

\begin{table}[hbt]
\begin{center}
\begin{tabular}{ c| c| c |c}
  \hline
  Type &Random rotation &  \multicolumn{2}{c}{Fraction of occluded area}\\
  \hline
  Dataset &All &  LineMOD &  LineMOD Occlusion, T-Less \\
 \hline
   Range &$\mathcal{U}$(-45$^\circ$, -45$^\circ$) & $\mathcal{U}$(0, 0.1)&$\mathcal{U}$(0.04, 0.5)\\
  \hline
\end{tabular}
\end{center}
\caption{Occlusion and rotation augmentation}
\end{table}

\subsection{The pools of symmetric poses for the transformer loss}
$I$: Identity matrix, $R^{\Theta}_a$: Rotation matrix about the $a$-axis with an angle $\Theta$. 
\begin{itemize}
    \setlength\itemsep{0.2em}
    \item{LineMOD and LineMOD Occlusion - eggbox and glue:  $sym=[I,R^{\pi}_z]$ }
    \item{T-Less - obj-5,6,7,8,9,10,11,12,25,26,28,29:  $sym=[I,R^{\pi}_z]$ }
    \item{T-Less - obj-19,20: $sym=[I,R^{\pi}_y]$ }
    \item{T-Less - obj-27: $sym=[I,R^{\frac{\pi}{2}}_z,R^{\pi}_z,R^{\frac{3\pi}{2}}_z]$ }
    \item{T-Less - obj-1,2,3,4,13,14,15,16,17,18,24,30: $sym=[I]$, the $z$-component of the rotation matrix is ignored.}
    \item{Objects not in the list (non-symmetric): $sym=[I]$}
\end{itemize}

\subsection{Pose prediction}
\begin{itemize}
    \setlength\itemsep{0.1em}
\item{Definition of non-zero pixels: $||I_\textrm{3D}||_2 >0.3$, $I_\textrm{3D}$ in normalized coordinates.}
\item{PnP and RANSAC algorithm: the implementation in OpenCV 3.4.0~\cite{opencv_library} is used with default parameters except the re-projection threshold $\theta_{re}$= 3.}
\item{List of outlier thresholds
\setlength{\tabcolsep}{4.5pt}
\begin{table}[hbt] 
\begin{center}
\begin{tabular}{ c| c c c c c c c c c c c c c}
  \hline
 \space &ape & bvise & cam&can & cat & driller & duck & eggbox & glue & holep & iron & lamp & phone\\
 \hline
 $\theta_o$ & 0.1 & 0.2 & 0.2 & 0.2 & 0.2 & 0.2 & 0.1 & 0.2 &0.2 &0.2 & 0.2 & 0.2 & 0.2 \\
  \hline

\end{tabular}
\end{center}
\caption{Outlier thresholds $\theta_o$ for objects in LineMOD}
\end{table}

\setlength{\tabcolsep}{6pt}
\begin{table}[hbt] 
\begin{center}
\begin{tabular}{ c| c c c c c c c c}
  \hline
\space &ape &can & cat & driller & duck & eggbox & glue & holep\\
 \hline
 $\theta_o$ & 0.2 & 0.3 & 0.3& 0.3 & 0.2 &0.2 & 0.3 & 0.3 \\
 \hline

\end{tabular}
\end{center}
\caption{Outlier thresholds $\theta_o$ for objects in LineMOD Occlusion}
\end{table}

\setlength{\tabcolsep}{1.7pt}
\begin{table}[hbt] 
\begin{center}
\begin{tabular}{ c| c c c c c c c c c c c c c c c c c c c c c c c c c c c c c c}
  \hline
 \space &01 &02 & 03 & 04 & 05 & 06 & 07 & 08 & 09 & 10 & 11 & 12 & 13 & 14 & 15 & 16 & 17&18&19&20&21&22&23&24&25&26&27&28&29&30\\
 \hline
 $\theta_o$  &0.1 &0.1& 0.1 & 0.3 & 0.2 & 0.3 &0.3 & 0.3 & 0.3 &0.2 & 0.3 & 0.3 & 0.2 & 0.2 & 0.2 & 0.3 & 0.3&0.2&0.3&0.3&0.2&0.2&0.3&0.1&0.3&0.3&0.3&0.3&0.3&0.3 \\
 \hline

\end{tabular}
\end{center}
\caption{Outlier thresholds $\theta_o$ for objects in T-Less}
\end{table}

\begin{figure}[hbt]
\begin{center}
   \includegraphics[width=0.75\linewidth]{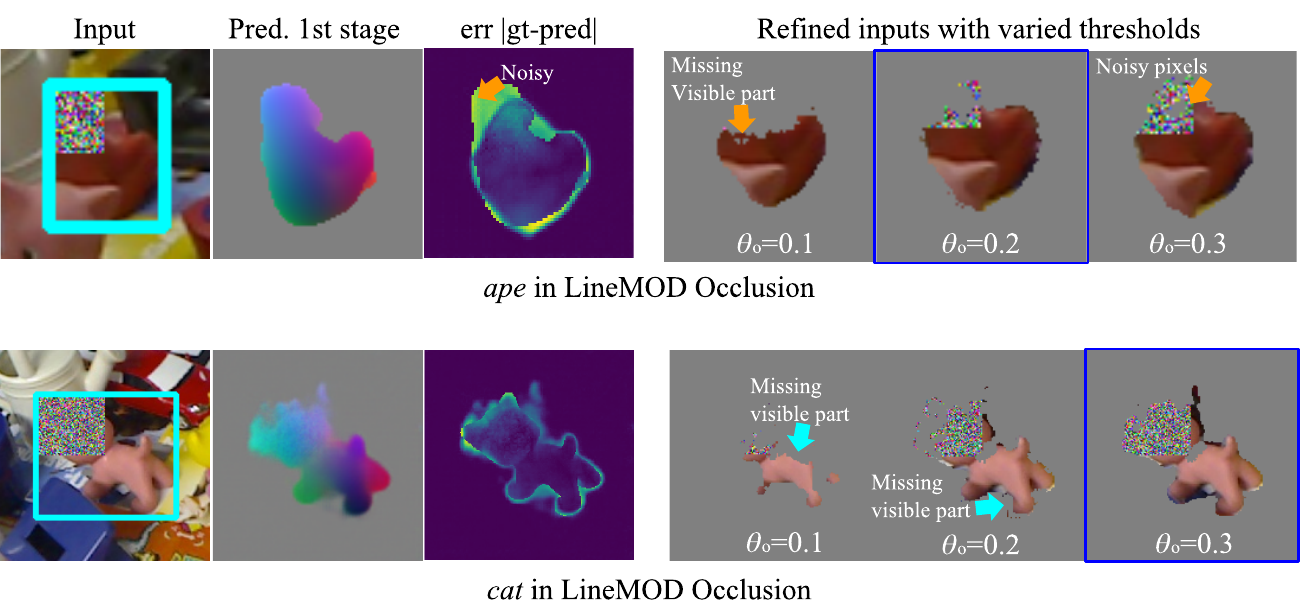}
\end{center}
   \caption{Examples of refined inputs in the first stage with varied values for the outlier threshold. Values are determined to maximize the number of visible pixels while excluding noisy predictions in refined inputs. Training images are used with artificial occlusions. The brighter pixel in images of the third column represents the larger error.}
\label{fig:outlier}
\end{figure}
}
\end{itemize}

\newpage
\section{Details of evaluations}
\vspace{5pt}

\subsection{T-Less: Object-wise results}
\setlength{\tabcolsep}{12pt}

\begin{table}[hbt]\centering 
\begin{center}
\begin{tabular}{ c| c c c c c c c c c c}
  \hline
 Obj.No &01 &02 & 03 & 04 & 05 & 06 & 07 & 08 & 09 & 10  \\
 \hline
 VSD Recall   &38.4 &35.3& 40.9 & 26.3 & 55.2 & 31.5 &1.1 & 13.1 & 33.9 &45.8  \\
  \hline
  \hline
  Obj.No& 11 & 12 & 13 & 14 & 15 &16 & 17 & 18 & 19 & 20  \\
 \hline
 VSD Recall & 30.7 & 30.4 &31.0 & 19.5 & 56.1  &66.5&37.9 &45.3& 21.7 & 1.9   \\
 \hline
 \hline
  Obj.No & 21 & 22 & 23 & 24 & 25 &26 & 27 & 28 & 29& 30 \\
 \hline
 VSD Recall&19.4 & 9.5 &30.7 & 18.3 & 9.5 &13.9 & 24.4 & 43.0 & 25.8 & 28.8     \\
 \hline
 
 \end{tabular}
 \vspace{+1em}
\caption{Object reall ($e_\textrm{vsd} < 0.3, \tau=20mm, \delta=15mm$) on all test scenes of Primesense in T-Less. Objects visible more than 10\% are considered. The bounding box of an object with the highest score is used for estimation in order to follow the test protocol of 6D pose benchmark~\cite{Hodan_2018_ECCV_bop}.}
\label{table:t-less}
\end{center}
\end{table}

\newpage
\subsection{Qualitative examples of the transformer loss}
\begin{figure}[hbt]
\begin{center}

   \def\svgwidth{\linewidth}
    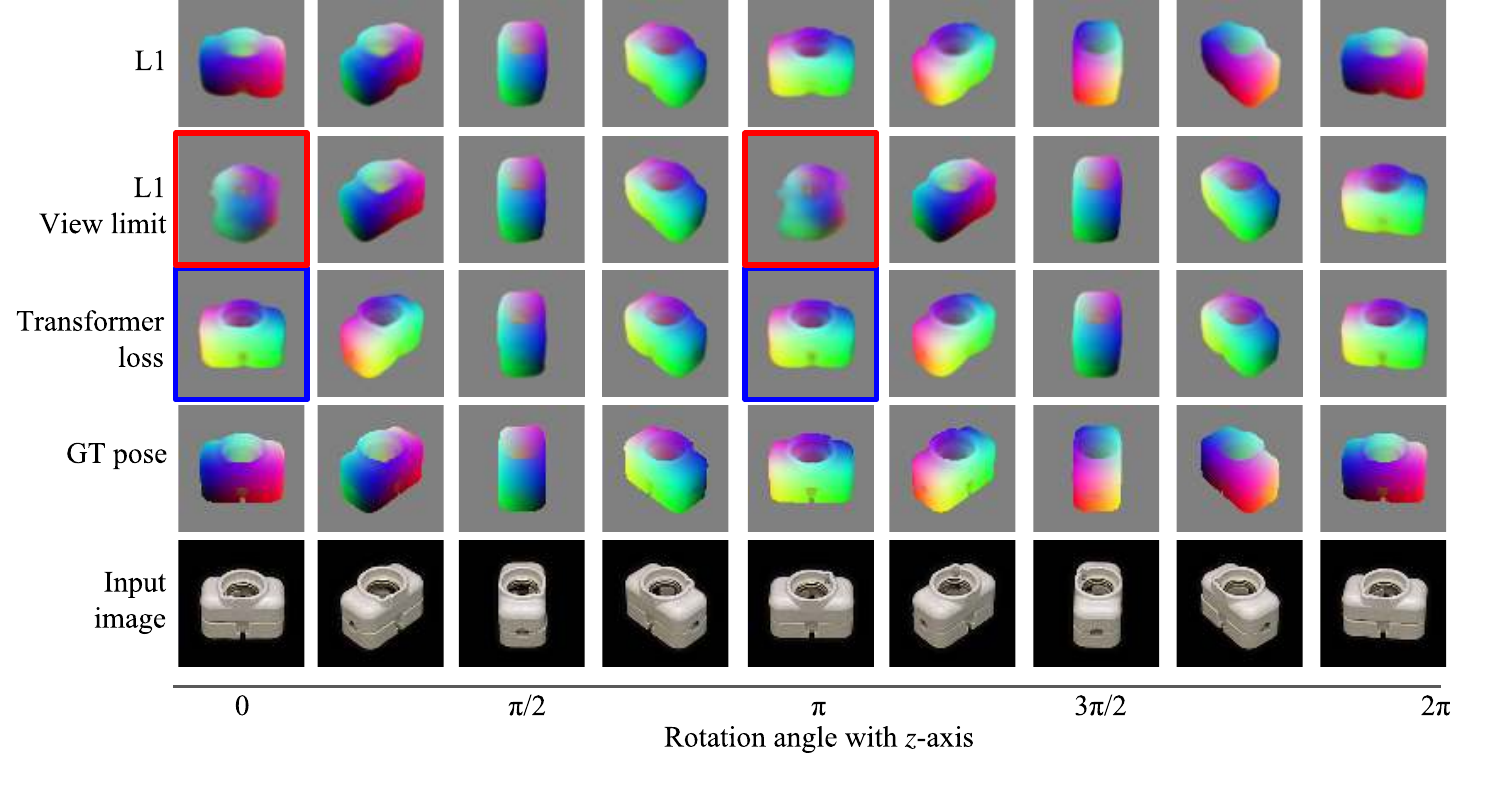      	
\end{center}
   \vspace{-15pt}
   \caption{Prediction results of varied rotations with the $z$-axis. As discussed in the paper, limiting a view range causes noisy predictions at boundaries, 0 and $\pi$, as denoted with red boxes. The transformer loss implicitly guides the network to predict a single side consistently. For the network trained by the L1 loss, the prediction is accurate when the object is fully visible. This is because the upper part of the object provides a hint for a pose.}
\label{fig:loss1}
\end{figure}

\begin{figure}[hbt]
\begin{center}
   \includegraphics[width=0.9\linewidth]{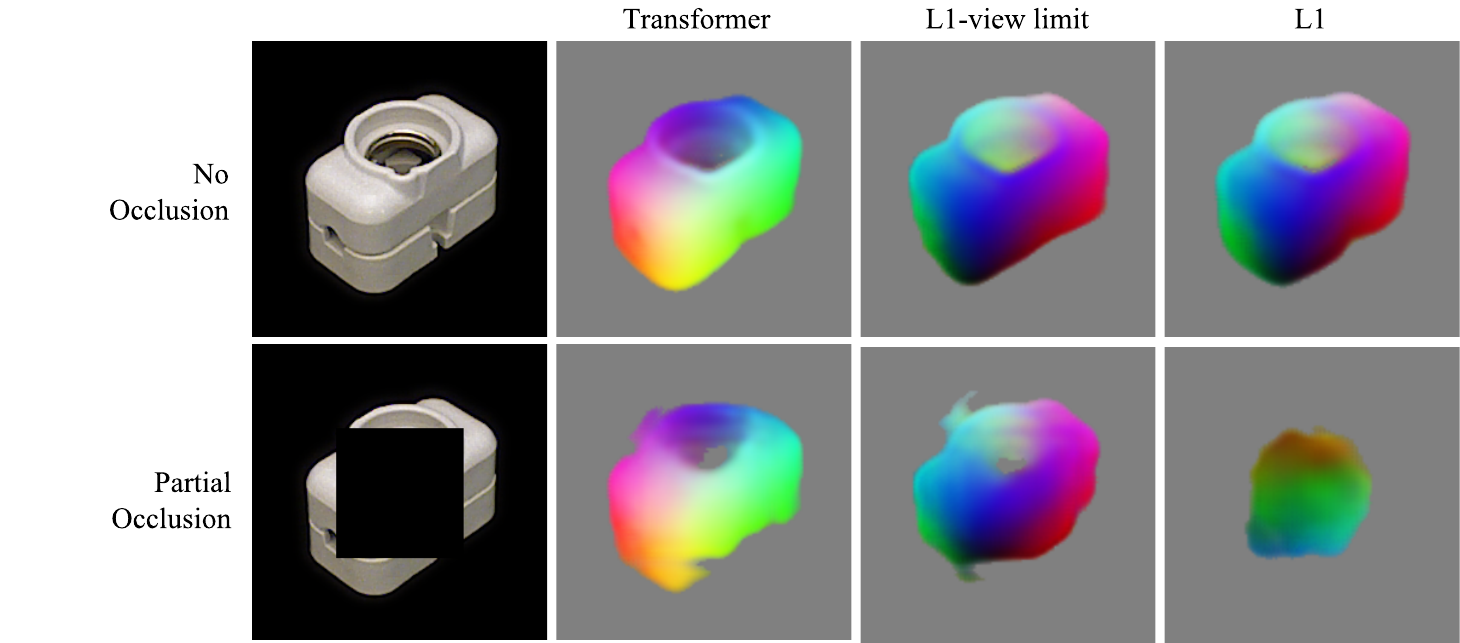}
\end{center}
  \vspace{-5pt}
  \caption{Prediction results with/without occlusion. For the network trained by the L1 loss, it is difficult to predict the exact pose when the upper part, which is a clue to determine the pose, is not visible. The prediction of the network using the transformer loss is robust to this occlusion since the network consistently predicts a single side.}
\label{fig:loss2}
\end{figure}

Figure~\ref{fig:loss1} and Figure~\ref{fig:loss2} present example outputs of the Pix2Pose network after training of the network with/without using the transformer loss. The \textit{obj-05} in T-less is used. 

\newpage
\vspace{-30pt}
\subsection{Example results on LineMOD}
\begin{figure}[hbt]
\begin{center}
   \def\svgwidth{\linewidth}
    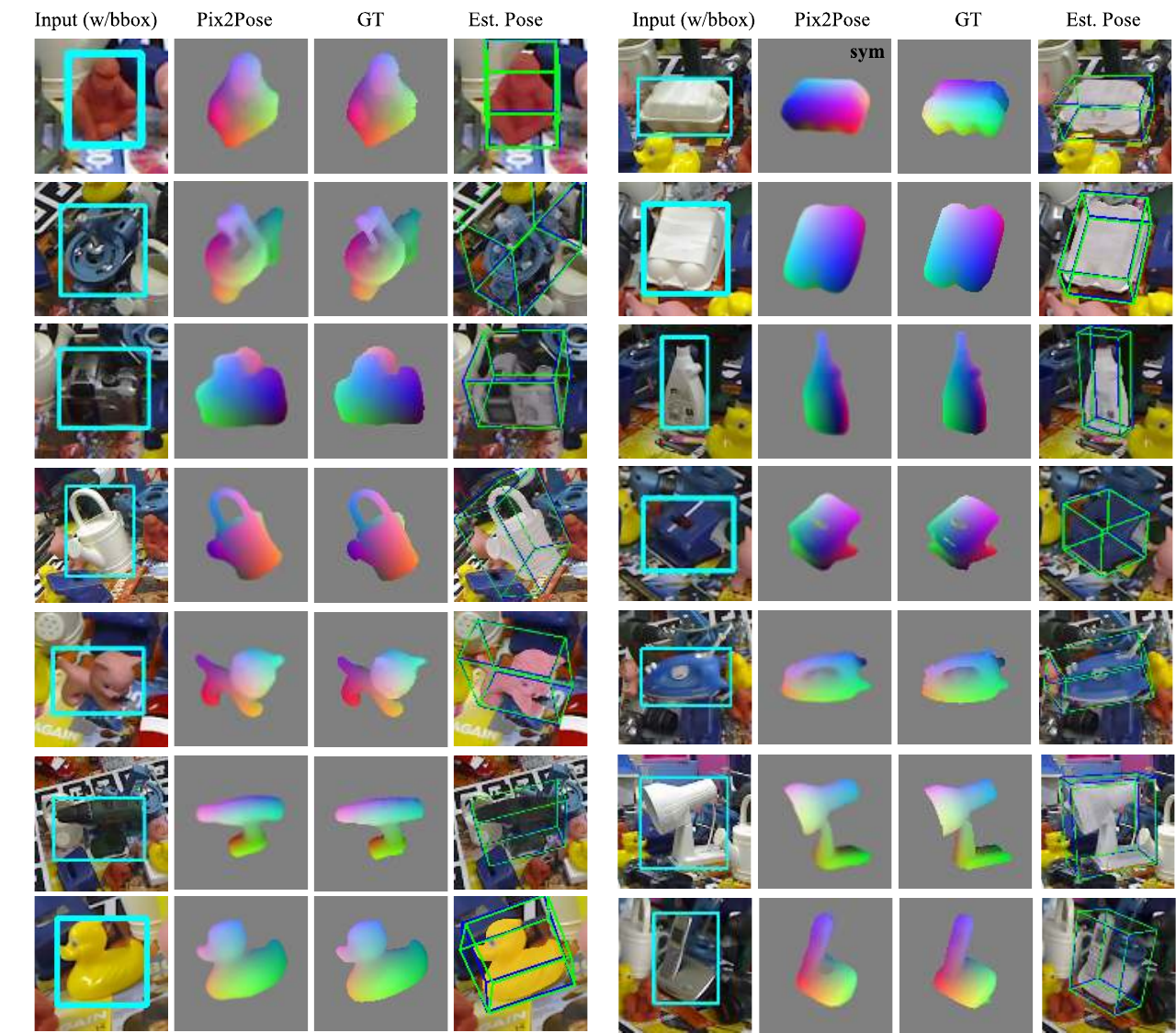      	
\end{center}
   \caption{Example results on LineMOD. The result marked with \textit{sym} represents that the prediction is the symmetric pose of the ground truth pose, which shows the effect of the proposed transformer loss. Green: 3D bounding boxes of ground truth poses, blue: 3D bounding boxes of predicted poses.}
\label{fig:linemod_result}
\end{figure}

\pagebreak
\subsection{Example results on LineMOD Occlusion} 
\begin{figure}[hbt]
\begin{center}
   \def\svgwidth{\linewidth}
    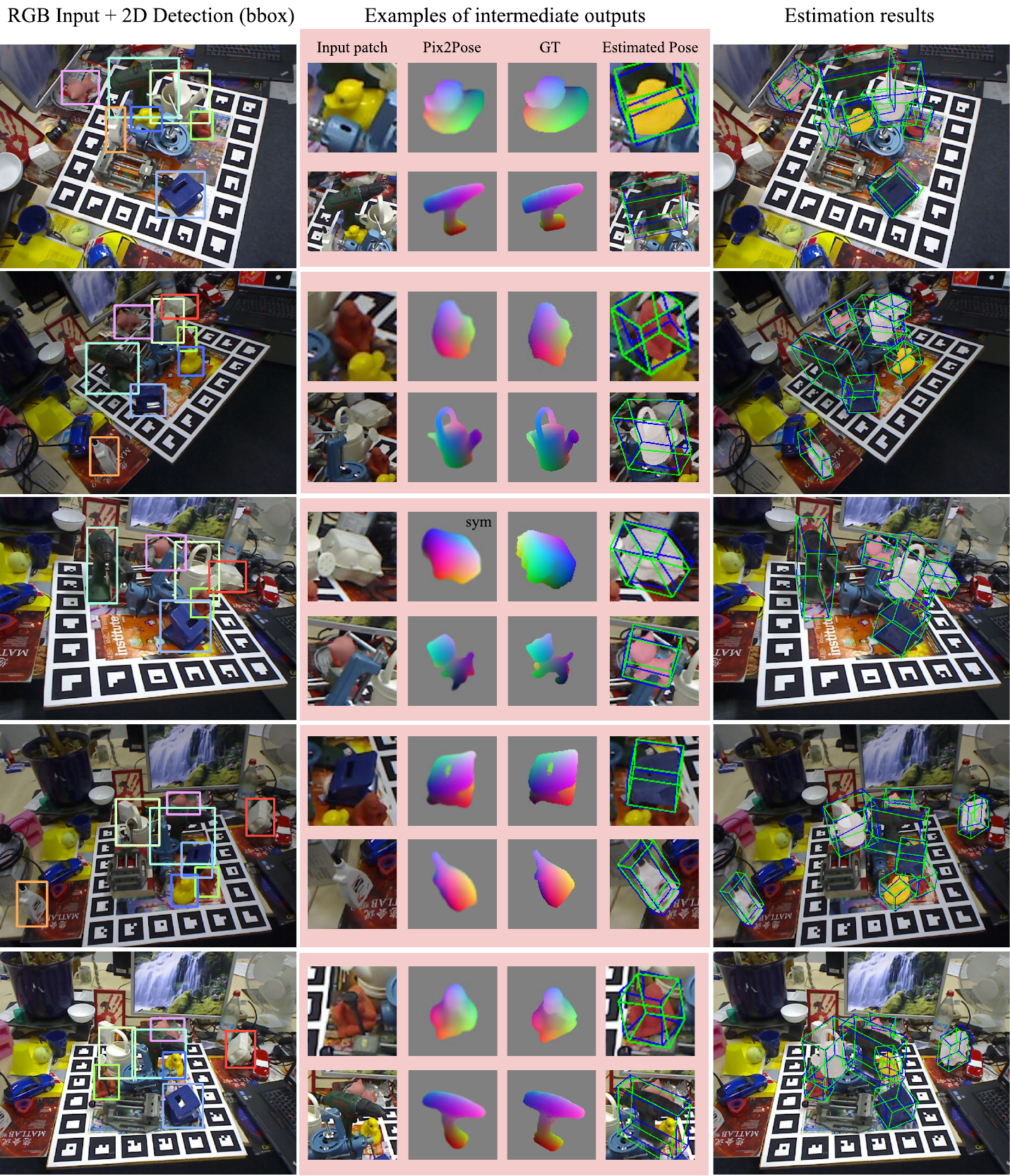      	
\end{center}
   \caption{Example results on LineMOD Occlusion. The precise prediction of occluded parts enhances robustness.}
\label{fig:occ_result}
\end{figure}
\pagebreak
\subsection{Example results on T-Less}

\begin{figure}[hbt]
\begin{center}
   \def\svgwidth{\linewidth}
   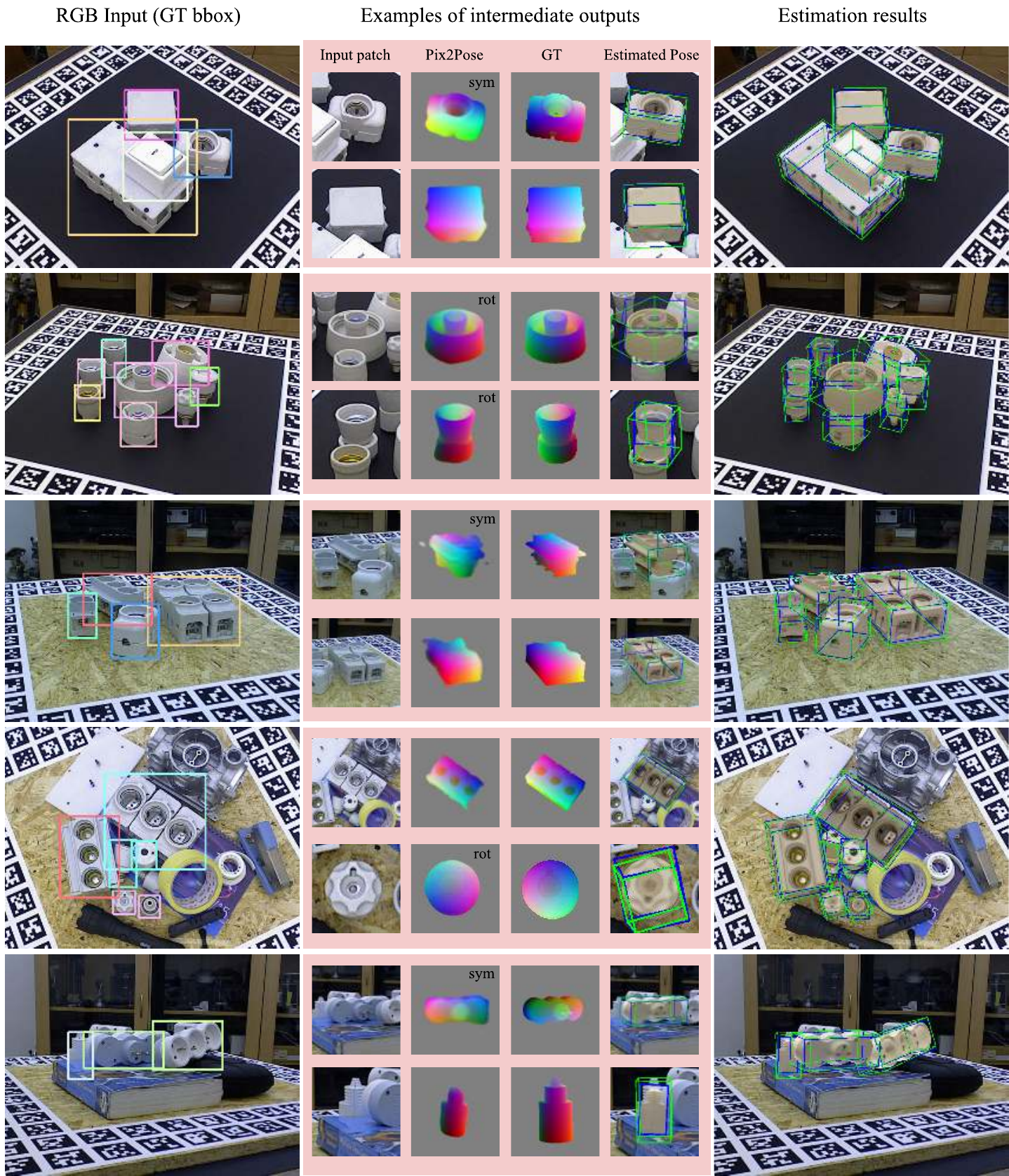      	
\end{center}
   \vspace{-3pt}
   \caption{Example results on T-Less. For visualization, ground-truth bounding boxes are used to show pose estimation results regardless of the 2D detection performance. Results with \textit{rot} denote estimations of objects with cylindrical shapes.}
\label{fig:tless_result}
\end{figure}

\section{Failure cases}
\begin{figure}[hbt]
\begin{center}
   \def\svgwidth{\linewidth}
    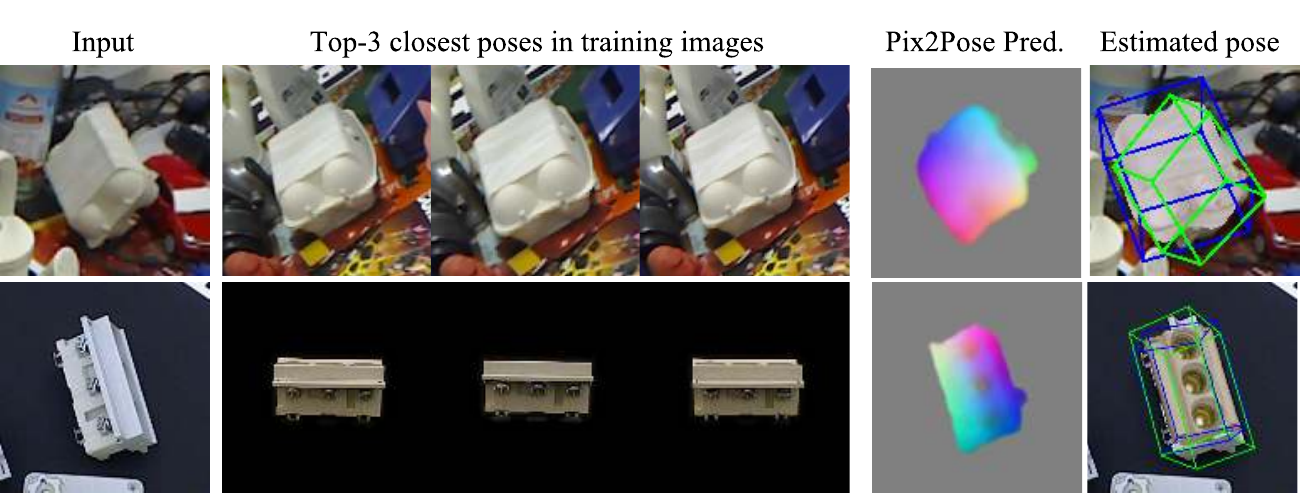      	
\end{center}
   \caption{Examples of failure cases due to unseen poses. The closest poses are obtained from training images using geodesic distances between two transformations (rotation only). 
   }
\label{fig:failure}
\end{figure}

Primary reasons of failure cases: (1) Poses that are not covered by real training images and the augmentation. (2) Ambiguous poses due to severe occlusion. (3) Not sufficiently overlapped bounding boxes, which cannot be recovered by the bounding box adjustment in the first stage. The second row of Fig.~\ref{fig:failure} shows that the random augmentation of in-plane rotation during the training is not sufficient to cover various poses. Thus, the uniform augmentation of in-plane rotation has to performed for further improvement.

\end{document}